\documentclass[runningheads]{llncs}
\usepackage{graphicx}

\usepackage{tikz}
\usepackage{comment}
\usepackage{amsmath,amssymb} 
\usepackage{color}
\usepackage{multirow}

\usepackage[hidelinks,colorlinks=true,linkcolor=red,citecolor=cyan,urlcolor=magenta]{hyperref}

\usepackage[accsupp]{axessibility}  

\usepackage{orcidlink}

\begin{document}
\pagestyle{headings}
\mainmatter
\def\ECCVSubNumber{xxxx}  

\title{Pointly-Supervised Panoptic Segmentation} 

\titlerunning{Pointly-Supervised Panoptic Segmentation}
%
\author{Junsong Fan\inst{1,3}\orcidlink{0000-0001-6989-2711} \and
Zhaoxiang Zhang*\inst{1,2,3}\orcidlink{0000-0003-2648-3875} \and
Tieniu Tan\inst{1,2}\orcidlink{0000-0002-1808-2169}}

\authorrunning{J. Fan et al.}
%
\institute{
Center for Research on Intelligent Perception and Computing, \\ Institute of Automation, Chinese Academy of Sciences, Beijing, China \and
University of Chinese Academy of Sciences, Beijing, China \and
Centre for Artificial Intelligence and Robotics, HKISI\_CAS, HongKong, China \\
\email{\{fanjunsong2016@,zhaoxiang.zhang@,tnt@nlpr.\}ia.ac.cn}}
\maketitle

\begin{abstract} 
In this paper, we propose a new approach to applying point-level annotations for weakly-supervised panoptic segmentation. Instead of the dense pixel-level labels used by fully supervised methods, point-level labels only provide a single point for each target as supervision, significantly reducing the annotation burden. We formulate the problem in an end-to-end framework by simultaneously generating panoptic pseudo-masks from point-level labels and learning from them. To tackle the core challenge, i.e., panoptic pseudo-mask generation, we propose a principled approach to parsing pixels by minimizing pixel-to-point traversing costs, which model semantic similarity, low-level texture cues, and high-level manifold knowledge to discriminate panoptic targets. We conduct experiments on the Pascal VOC and the MS COCO datasets to demonstrate the approach's effectiveness and show state-of-the-art performance in the weakly-supervised panoptic segmentation problem. Codes are available at \url{https://github.com/BraveGroup/PSPS.git}.

\keywords{weakly-supervised learning, panoptic segmentation}
\end{abstract}

\section{Introduction}
\label{sec:introduction}

Panoptic segmentation~\cite{panoptic_segmentation} aims at fully parsing all the pixels into nonoverlapping masks for both thing instances and stuff classes. It combines the semantic segmentation and the instance segmentation tasks simultaneously. Classical deep learning approaches require precise dense pixel-level labels to solve this problem. However, acquiring exact pixel- and instance-level annotations on large-scale datasets is very time-consuming, hindering the popularization and generalization of the approaches in new practical applications.

To alleviate the annotation burden for segmentation models, researchers recently proposed weakly-supervised learning (WSL)~\cite{survey:wsss,survey:wsss2,survey:wsod}, which focuses on leveraging coarse labels to train dense pixel-level segmentation tasks. Typically, the weak supervision includes image-level~\cite{wsss:icd,wsss:cian,wsss:multi_seed}, point-level~\cite{wsss:whats_point,wsss:pdml}, scribble-level~\cite{wsss:scribblerw,wsss:scribblesup}, and bounding box-level labels~\cite{wsss:boxsup}, etc.
These approaches tackle either semantic segmentation~\cite{wsss:ccnn}, instance segmentation~\cite{wsss:irnet,wsss:sdi_box}, or panoptic segmentation~\cite{wsss:jtsm,wsss:wsps} tasks.
Among them, the weakly-supervised panoptic segmentation (WSPS) problem is the most challenging since it requires both semantic and instance discrimination with only weak supervision.
As a result, the WSPS got less attention in previous works, and its performance is far from satisfactory. The seminal work by Li et al.~\cite{wsss:wsps} manages to address the WSPS problem using bounding-box level labels. Later, JTSM~\cite{wsss:jtsm} proposes to apply only image-level labels for the WSPS problem. Recently, PanopticFCN~\cite{panoptic_fcn} tackles this problem by connecting multiple point labels into polygons. The performances of these approaches differ significantly with the different weak annotations. 

\begin{figure}[t]
\centering
\includegraphics[width=0.98\textwidth]{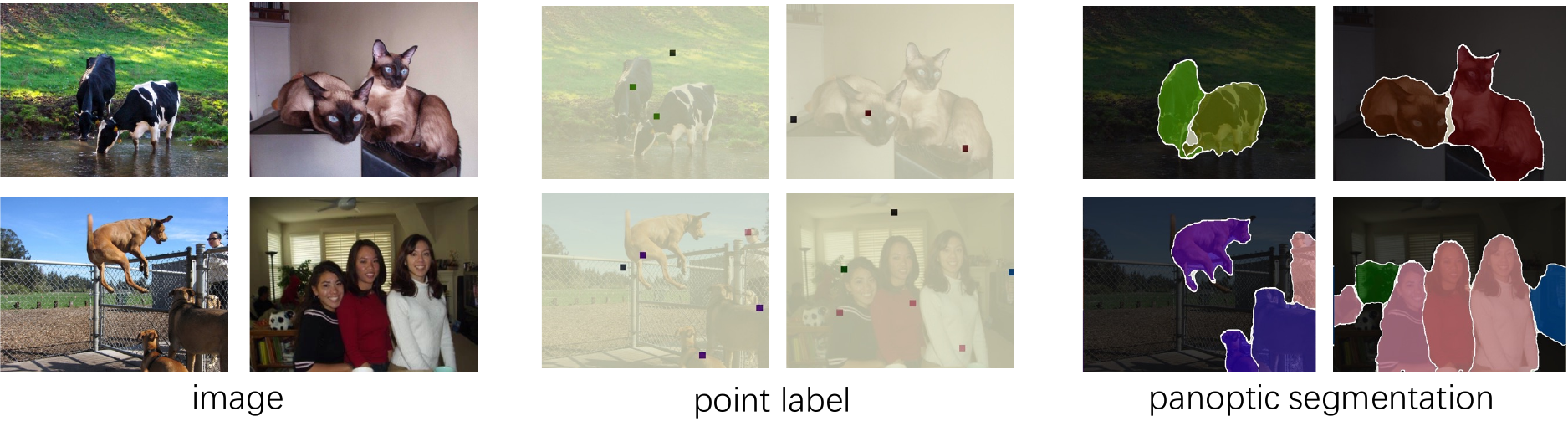}
\caption{Illustration of the proposed pointly-supervised panoptic segmentation. From left to right: input images, point labels as supervision, and panoptic segmentation predictions. The point labels provide a single point annotation for each target, including both thing instances and stuff classes, which are used at training time only. Please see Sec. \ref{sec:approach} for details. Best viewed in color.}
\label{fig:setting}
\end{figure}

In this paper, we propose a new WSPS paradigm to use only a single point for each target as the supervision, as illustrated in Fig.~\ref{fig:setting}. Recall that the core of weakly-supervised segmentation is to release the annotation burden while still obtaining decent performance. In other words, balance the cost of annotation and the model performance. We are motivated to use the point-level labels because, on the one hand, the annotation time of point-level labels is only marginally above the image-level labels~\cite{wsss:whats_point}, saving much cost compared with the box-level or polygon labels. On the other hand, point labels can provide minimum spatial information to localize and discriminate different panoptic targets for the segmentation models.

A natural idea to estimate panoptic masks from point-level labels is to assign each pixel in the image to one of the points according to some principles. To this end, we propose tackling this problem by minimizing the pixel-to-point traversing cost, measured by the neighboring pixel affinities. There are two basic requirements to correctly assign pixels to point labels: semantic class matching and instance discrimination. The former ensures that the pixels are parsed with the correct class labels, and the latter is responsible for distinguishing different instances in the thing classes. Therefore, we consider three criteria to model the affinities: semantic similarity, low-level image cues, and high-level instance discrimination knowledge. Using these criteria, we model the pixel-to-point traversing costs and solve the assignment problem by finding the shortest path.

We base our approach on the transformer models~\cite{transformer,bert,gpt}, which have recently shown impressive results in computer vision tasks~\cite{maskformer,panoptic_segformer,segmenter,detr,deformable_detr}. Specifically, our approach contains a group of semantic query tokens to parse semantic segmentation results and a group of panoptic query tokens responsible for the panoptic segmentation task~\cite{panoptic_segformer}.
In addition to the regular panoptic segmentation model, our approach contains a label generation model, which produces dense panoptic pseudo-masks depending on the point-level labels and the criteria above.
The whole approach is end-to-end. After training, only the panoptic segmentation branch is kept for testing. Thus, it does not incur additional computation or memory overhead for usage. We conduct thorough experiments to analyze the proposed approach and the properties of the point-level labels. Meanwhile, we demonstrate new cutting-edge performance with the WSPS problem on the Pascal VOC~\cite{data:voc} and the MS COCO~\cite{data:coco} datasets.

In summary, the main contributions of this work are:

\begin{itemize}
	\item We propose a new paradigm for the WSPS problem, which utilizes a single point for each target as supervision for training.
	\item A novel approach to estimating dense panoptic pseudo-masks by minimizing the pixel-to-point traversing distance is proposed.
	\item We implement the approach in an end-to-end framework with transformers, conduct analytical experiments to study the model and the point-level labels, and demonstrate state-of-the-art performance on the Pascal VOC and the MS COCO datasets.
\end{itemize}

\section{Related Works}
\label{sec:related_work}

\subsection{Panoptic Segmentation}

The panoptic segmentation~\cite{panoptic_segmentation} task simultaneously incorporates semantic segmentation and instance segmentation, where each pixel is uniquely assigned to one of the stuff classes or one of the thing instances.
This problem can be tackled by combining the semantic and instance segmentation results in a post-processing manner~\cite{panoptic_segmentation}. Later works such as JSIS~\cite{jsis} adopt a unified network combining a semantic segmentation branch and an instance segmentation branch. After that, many approaches have been proposed for improvement by using feature pyramids~\cite{panoptic_fpn}, automatic architecture searching~\cite{auto_panoptic}, and unifying the pipeline~\cite{unify_panoptic}, etc.

Recently, transformer-based approaches have shown impressive results across the NLP~\cite{transformer,bert,gpt} and the CV~\cite{detr,vit,deformable_detr,swin} applications. The seminal work DETR~\cite{detr} provides a clear and elegant solution for object detection and segmentation. The following work DeformableDETR~\cite{deformable_detr} improves it by using the deformable transformers to reduce the computation burden and accelerate the convergence. K-Net~\cite{knet} adopts an iterative refinement procedure to enhance the attention masks gradually. MaskFormer~\cite{maskformer} proposes to separate the mask prediction and the classification process. Panoptic SegFormer~\cite{panoptic_segformer} embraces a similar idea and adopts an auxiliary localization target to ease the model training. Our panoptic segmentation approach is based on these works, and we focus on alleviating the annotation burden by exploiting point-level annotations.

\subsection{Weakly-Supervised Segmentation}

Weakly supervised segmentation~\cite{survey:wsss,survey:wsss2} aims to alleviate the annotation burden for segmentation tasks by using weak labels for training.
According to the type of tasks, it concerns semantic segmentation~\cite{wsss:cian,wsss:ccnn,wsss:sec,wsss:mdc,wsss:ficklenet,wsss:aisi}, instance segmentation~\cite{wsss:irnet,wsss:boxinst}, and panoptic segmentation~\cite{wsss:wsps,wsss:jtsm} problems. According to the kinds of supervision, these approaches use image-level~\cite{wsss:cian,wsss:ccnn,wsss:sec,wsss:mdc,wsss:ficklenet,wsss:jtsm}, point-level~\cite{wsss:whats_point,wsss:pdml}, scribble-level~\cite{wsss:scribblerw,wsss:scribblesup}, or box-level~\cite{wsss:frgbox,wsss:boxinst,wsss:boxsup} labels for training. Among them, image-level label-based approaches are the most prevalent. These approaches generally rely on the CAM~\cite{wsss:cam,wsss:grad_cam} to extract spatial information from classification models, which are trained by the image-level labels. Though great progress has been achieved by these approaches on the semantic segmentation task, it is generally hard to distinguish different instances of the same class with only image-level labels, especially on large-scale datasets with many overlapping instances. Li et al.~\cite{wsss:wsps} proposes to address this problem by additionally using bounding-box annotations, which however takes much more time to annotate. PanopticFCN~\cite{panoptic_fcn} alternatively proposes to use coarse polygons to supervise the panoptic segmentation model, which are obtained by connecting multiple point annotations for each target. PSIS~\cite{wsss:psis} proposes to address the instance segmentation problem by using sparsely sampled foreground and background points in each bounding box.
Though these approaches achieve better results, their annotation burden is significantly heavier than image-level labels. In this paper, we try to use a new form of weak annotation for panoptic segmentation, i.e., a single point for each target. We demonstrate that this supervision can achieve competitive performance compared with previous approaches while significantly reducing the annotation burden.

\subsection{Point-Level Labels in Visual Tasks}

Recently point-level annotation has drawn interest in a broad range of computer vision tasks. Beside the works concerning the detection and segmentation tasks~\cite{wsss:psis,panoptic_fcn,wsss:pdml,wsss:whats_point}, some works adopt point-level labels to train crowd counting~\cite{crowd_counting,crowd_counting2} models. SPTS~\cite{peng2021spts} proposes to use points for the text spotting problem. Chen et al.~\cite{chen2021points} propose addressing weakly-supervised detection problems using point labels. Besides, point labels also play an essential role in interactive segmentation models, where users provide interactive hints through point-level clicks~\cite{iog,deep_extreme_cut,iterative_iterative}. To the best of our knowledge, there are still no approaches to training panoptic segmentation models using only a single point per target.

\section{Approach}
\label{sec:approach}

\begin{figure}[t]
\centering
\includegraphics[width=\textwidth]{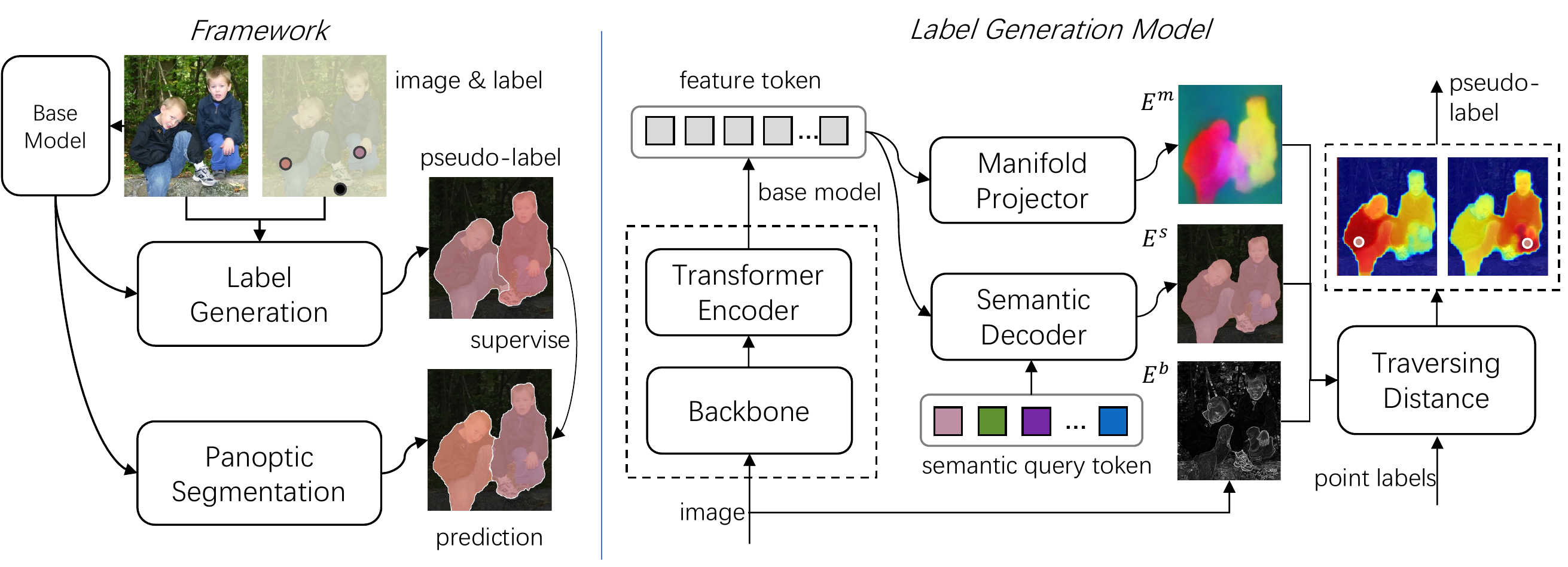}
\caption{Illustration of the proposed approach. Left: the overall framework, which contains a label generation model and a panoptic segmentation model. The former produces dense panoptic pseudo-labels from point-level labels, the latter is responsible for the final panoptic segmentation prediction. Right: the detailed pipeline of the label generation model. Please see Sec.~\ref{sec:approach} for more information. Best viewed in color.}
\label{fig:framework}
\end{figure}

In this section, we elaborate on the details of the proposed approach. Fig.~\ref{fig:framework} illustrates the overall framework, which can be decomposed into two major components, a label generation model and a panoptic segmentation model. These two components share the same backbone and the transformer encoder~\cite{deformable_detr}. The label generation model is the core of the weakly-supervised learning, which is responsible for obtaining dense panoptic pseudo-masks from weak point-level labels. The panoptic segmentation model is the same as those fully-supervised ones and learns from the panoptic pseudo-masks. All these models are trained as a whole in an end-to-end manner. After the training stage, the label generation model can be removed, only leaving a standard panoptic segmentation model for usage. Hence, it does not bring any computation or memory overhead than other fully-supervised methods.

\subsection{Dense Semantics from Point Supervision}

Semantic parsing is the cornerstone of our approach to producing dense pixel-level pseudo-masks from sparse point-level labels. We decompose the panoptic pseudo-label generation problem into two steps: semantic parsing and instance discrimination. In the semantic parsing step, the semantic probabilities for all the pixels in the image are first obtained. By means of this, the latter problem could be reduced to partition pixels within each class into different instances, reducing the solution space and improving the estimation robustness.

To generate semantic segmentation results, we adopt a set of semantic query tokens, which has a one-to-one correspondence to the semantic classes, as shown in Fig. \ref{fig:framework}. The semantic decoder is made of transformer decoder layers following the Panoptic SegFormer \cite{panoptic_segformer}. It contains a mask branch to decode masks from tokens and a classification branch to decode the class probabilities. The semantic segmentation probabilities are then obtained by multiplying together the mask probabilities and the class probabilities.

Let $P\in \mathbb{R}^{HW\times C}$ denote the semantic segmentation probabilities of the $C$ classes. Given a set of $N$ point-level labels, it can be mapped to a set of $N^s$ labeled semantic pixels, $Y = \{(x_i, c_i)\}_{i=1}^{N^s}$, where $x_i$ and $c_i$ are the coordinate and class index of the $i$th pixel, respectively. The mapping could be implemented by coloring the surrounding pixels of each point-level label and applying the same geometric augmentations as the input image. The partial cross-entropy loss for semantic segmentation is the average on labeled pixels:

\begin{equation}
	\mathcal{L}_{sem} = -\frac{1}{N^s} \sum_{(x_i, c_i)\in Y}\log P_{x_i, c_i},
\label{eq:pce_loss}
\end{equation}

To supplement the sparse partial cross-entropy loss, inspired by \cite{wsss:boxinst}, we adopt the image texture-based constraints densely on all the pixels, a.k.a., color-prior loss.
Let $P_i,P_j \in\mathbb{R}^C$ denote the class probabilities of the $i$th and $j$th pixels, respectively. The color prior loss is defined as:

\begin{equation}
	\mathcal{L}_{col} = -\frac{1}{Z} \sum_{i=1}^{HW}\sum_{j\in\mathcal{N}(i)} A_{i,j}\log P_{i}^T P_{j},
\label{eq:color_prior}
\end{equation}
where $P_{i}^T P_{j}$ measures the predicted probability similarity of the pair, higher values indicate that the prediction of the pair tends to be the same class. $A_{i,j}$ is the color-prior affinity following~\cite{wsss:boxinst}, which is obtained by thresholding the pixel similarity computed in the LAB color space with threshold 0.3. $\mathcal{N}(i)$ is the set of neighbor pixel indices of $i$. $Z=\sum_{i=1}^{HW}\sum_{j\in\mathcal{N}(i)} A_{i, j}$ is the normalization factor. By optimizing Eq.~\ref{eq:color_prior}, neighboring pixels owning similar colors are encouraged to derive the same semantic prediction. Experiments in Sec.~\ref{sec:semantic_segmentation} demonstrate this strategy can effectively improve the mask quality.

\subsection{Traversing Distance and Mask Generation}
\label{sec:distance}

\begin{figure}[t]
\centering
\includegraphics[width=0.95\textwidth]{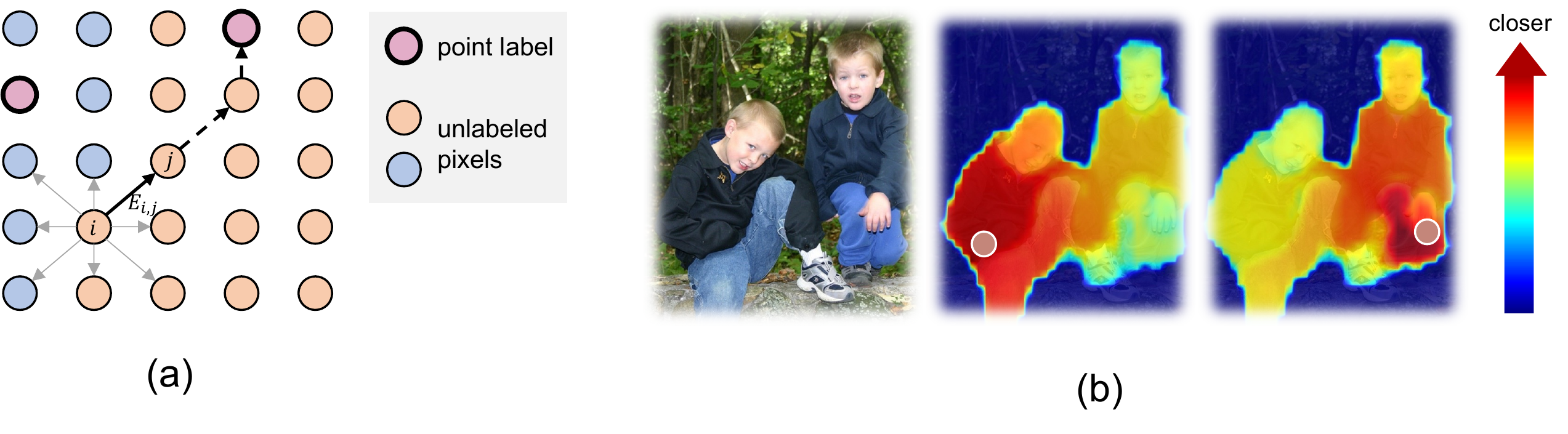}
\caption{(a) Illustration of the traversing distance method. Each pixel finds the point label with the minimum traversing cost and is assigned to the corresponding target. (b) Example of a two-instance case, showing the traversing cost of each pixel to each point label. Highlighted regions have low costs. Please see Sec.~\ref{sec:distance} for details. Best viewed in color.}
\label{fig:distance}
\end{figure}

After obtaining the semantic classes of each pixel, the challenge of generating panoptic masks is mainly reduced to discriminating different instances in the same class. We propose a principled approach to address this problem by assigning each pixel to the nearest point label, where the distance is defined by the proposed traversing distance, as illustrated in Fig.~\ref{fig:distance}.

Denote the cost of traveling from pixel $i$ to point label $s$ by $\mathcal{D}_{i, s}$, the target is to find the nearest point label $\hat{s}$, and mark pixel $i$ as the foreground of the corresponding segmentation target. Formally,

\begin{equation}
	\begin{aligned}
		\mathcal{D}_{i, s} &= \min_{\Gamma_{i,s}} \int E(x)\Gamma_{i,s}(x) dx	 \\
	\end{aligned},
\label{eq:assign_instance}
\end{equation}
where, $\Gamma_{i,s}$ is a path from pixel $i$ to point label $s$. $E$ describes the non-negative traversing cost along the path. In discrete digital images, the path is defined as a sequence of continual pixels under the 8-neighborhood connection.

In this framework, the question reduces to defining proper transferring cost $E$ to help distinguish different instances. We consider three criteria to accomplish this task: semantic similarity, low-level boundary cues, and high-level instance-aware manifold knowledge.
For clarity, we slightly abuse the notation to denote the cost between neighboring pixels $i$ and $j$ by $E_{i, j}$, which is composed of the three items:

\begin{equation}
	E_{i, j} = E_{i, j}^{s} + \lambda_b E_{i, j}^{b} + \lambda_m E_{i, j}^{m},
\label{eq:distance}
\end{equation}
where, $\lambda_b$ and $\lambda_m$ are the weights controlling the relative importance.

$E_{i, j}^{s}$ reflects the semantic similarity, whose value is low if the neighboring pixels belong to the same class. It is defined by the aforementioned semantic probabilities:

\begin{equation}
	E_{i, j}^s = \sum_{c=1}^C |P_{i, c} - P_{j, c}|,
\end{equation}
where, $P_{i,c}$ is the probability of the $i$th pixel belonging to class $c$. By means of this, crossing pixels of different classes are costly. Thus, paths residing in the object interiors are encouraged. This strategy could help pixels be assigned to the point within a coherent class region rather than geometrically closer points but with different classes.

$E_{i, j}^b$ defines the boundary cost considering the low-level image textures. Given the boundary map $B\in\mathbb{R}^{HW}$ obtained by edge filters, $E_{i, j}^b$ is counted by the non-negative value at the target location. In this way, the paths are encouraged not to cross the boundaries:

\begin{equation}
	E_{i, j}^b = |B_j|,
\label{eq:boundary}
\end{equation}

In this paper, we adopt the efficient Sobel filter to compute the boundary map. The boundary cost implicitly assumes that regions of coherent colors are more likely to belong to the same class and instance, which has been proved experimentally by previous works~\cite{wsss:sec,alg:crf,wsss:boxinst} in addressing the segmentation tasks.
 
$E_{i, j}^{m}$ provides high-level knowledge to distinguish instances, which is learned by the deep model online. We add a manifold projector to the transformer encoder to produce dense features to compute the instance similarity, as illustrated in Fig. \ref{fig:framework}. The manifold projector firstly reshapes the feature tokens back to 2D spatial features. Features from different pyramid levels are bilinearly sampled to the same size and summed together. Then, the projection is obtained by a 2-layer MLP model, implemented by two $1\times 1$ convolution layers interleaved by a ReLU activation. Given the L2-normalized feature map $F\in\mathbb{R}^{HW\times D}$ produced by the manifold projector, the cost is defined by the non-negative cosine distance:

\begin{equation}
	E_{i, j}^m = \max\{1 - F_i^TF_j, 0\},
\label{eq:projection}
\end{equation}
where, the projected features belonging to the same instance are similar and produce low costs. In this way, paths are encouraged not to cross different instances. To convey instance-aware knowledge, the manifold projection model should be trained by instance-aware constraints. For clarity, we postpone explaining the details in the Sec. \ref{sec:training_loss}.

After obtaining the criteria, the next step is finding the shortest path between each pair of pixels and point labels in the 8-neighborhood graph. Note that the graph is very sparse because each pixel only connects to its local neighbors. Let $N$ denote the total number of point labels and $M$ denote the number of pixels. Then, the minimum distance $\mathcal{D}_{i,s}$ in Eq. \ref{eq:assign_instance} can be efficiently solved by the shortest distance algorithm with time complexity $O(MN\log M)$.

It is noteworthy that the distance measurement in Eq. \ref{eq:assign_instance} can only produce connected components. As a result, if an instance is overlapped and separated into several parts by some region belonging to different classes, the farther parts would be assigned with the wrong class. To overcome this limitation, we use the class compatibility between the pixel and the point label to reweight the distance before assigning pixels to point labels.

\begin{equation}
	\hat{s} = \text{argmin}_{s}\left[\left(\hat{\mathcal{D}}_{i,s} - 1\right) \cdot \left(P_{i}^TP_{s}\right)\right],
\label{eq:reweight}
\end{equation}
where, $P_i$ is the probability of the $i$th pixel obtained by the semantic segmentation branch. $P_s$ is the one-hot encoding of the point label's ground truth class. $\hat{\mathcal{D}}$ is the normalized version of $\mathcal{D}$ in range $[0, 1]$.
Finally, we judge that pixel $i$ is part of the instance holding point label $\hat{s}$ according to Eq.~\ref{eq:reweight}, and the whole image is parsed into non-overlapping regions, obtaining panoptic pseudo-masks.

\subsection{Weakly-Supervised Training}
\label{sec:training_loss}

In this section, we elaborate on training the whole model with the generated pseudo-masks. Firstly, the panoptic segmentation model, as illustrated in Fig. \ref{fig:framework}, is trained by the pseudo-masks. As aforementioned, we adopt the Panoptic SegFormer architecture, which contains a localization decoder to help quickly converge to the target locations, a classification branch to predict class probabilities for each query token, and a mask decoder to decode masks. They are optimized by the localization loss, the focal loss, and the dice loss, respectively. For simplicity, here we denote all these losses to train the panoptic segmentation model as $\mathcal{L}_{pan}$. Please refer to the paper \cite{panoptic_segformer} for more details.

In addition to the panoptic segmentation model, the manifold projector used in Eq. \ref{eq:projection} also needs to train to provide instance-aware representations. Here we utilize a contrastive learning strategy \cite{selfsup:moco,selfsup:simclr} to optimize the manifolds with the pseudo-masks. Let $\mathcal{M}_i\in \mathbb{R}^{HW}$ denote the pseudo-mask of target $i$ that contains point label with coordinate $x_i$. The global representation of target $i$ is the masked average of the projection $F\in\mathbb{R}^{HW\times D}$:

\begin{equation}
	\bar{F}_{i} = \frac{1}{Z_i}\sum_{j=1}^{HW}\mathcal{M}_{i,j}F_{j},
\end{equation}
where, $Z_i = \sum_{j=1}^{HW}\mathcal{M}_{i,j}$ is for normalization, $F_j\in\mathbb{R}^D$ is the feature at pixel $j$.

The loss for the projector is the average of all point-to-target contrasts:

\begin{equation}
	\mathcal{L}_{cl} = -\frac{1}{N}\sum_{i=1}^N \log \frac{\exp\left( F_{x_i}^T \bar{F}_i /\tau \right)}{\sum_{j=1}^N \exp\left( F_{x_i}^T \bar{F}_j /\tau \right)},
\end{equation}
where, $N$ is the total number of point labels, $\tau$ is a scale factor and is set 0.07 following \cite{selfsup:moco}. When the setting is extended to annotate each target with multiple points, the sum in the denominator of the contrasts should iterate over the number of targets.
With this optimization target, the feature projections at the labeled points are encouraged to be coherent within the estimated instance region and distinctive from the others. The pseudo-mask estimation and the manifold projector mutually benefit each other and improve the model together. 

Taking all the above components together, our final model is end-to-end:

\begin{equation}
	\mathcal{L}_{all} = \mathcal{L}_{sem} + \mathcal{L}_{col} + \mathcal{L}_{pan} + \mathcal{L}_{cl},
\end{equation}

After training, only the panoptic segmentation model is kept for testing, thus not incurring any computation or memory overhead compared with previous fully-supervised panoptic segmentation models.

\section{Experiments}
In this section, we discuss the experiments. We first describe the experiment setting and implementation details and then elaborate on the analyses and comparison using the VOC and COCO datasets.

\subsection{Datasets}

\noindent\textbf{Pascal VOC} dataset~\cite{data:voc,data:voc_extra} contains 10582 training images and 1449 validation images. It has 20 thing classes and one stuff class. By default, a single point label per target is sampled with the uniform distribution from the masks, which is fixed through all the experiments. To study the influence of point label distribution, we also adopt other sampling strategies in Sec.~\ref{sec:sampling_strategy} for analyses. We analyze our approach with the panoptic quality (PQ), segmentation quality (SQ), recognition quality (RQ), and intersection over union (IoU) metrics. 

\noindent\textbf{MS COCO} dataset~\cite{data:coco} contains 118k images for training, 5k images for validation, and 20k images for testing. It contains 80 thing classes and 53 stuff classes. The same point sampling strategy and metrics are applied to the COCO.

\subsection{Implementation Details}

\textbf{Architecture.} We base our approach on the Panoptic SegFormer with a ResNet50 backbone~\cite{net:resnet_v1}. As mentioned in Sec.~\ref{sec:semantic_segmentation}, we adopt an extra group of semantic query tokens to produce the semantic segmentation results. Specifically, the $C$ semantic tokens produce $C$ masks and classification scores, which are multiplied together and projected by a linear layer followed by a Softmax function to produce semantic probabilities. The mask decoder contains 6 transformer decoder layers~\cite{panoptic_segformer}, which has the same architecture as the panoptic segmentation model. The color prior loss in Eq.~\ref{eq:color_prior} is constructed by sampling neighboring pixels with kernel size 5 and dilation rate 2, and the loss is amplified by factor 3. The image edge used in Eq.~\ref{eq:boundary} is obtained by the Sobel filter in the LAB color space and its absolute values are normalized into the range $[0, 1]$. The panoptic segmentation model follows the same setting as~\cite{panoptic_segformer}, and the number of query tokens for panoptic segmentation is set to 300.

~

\noindent\textbf{Optimization.} We follow previous practice~\cite{panoptic_segformer} for training, i.e., AdamW optimizer with learning rate $1.4\times 10^{-4}$, weight decay $1.4\times 10^{-4}$, and batch size 8. The learning rates for the backbone parameters are multiplied by the factor $0.1$. To stabilize the training, we adopt a linear warm-up strategy for the losses $\mathcal{L}_{pan}$ and $\mathcal{L}_{cl}$ during the first training epoch, so that reliable pseudo-panoptic masks can be obtained, which depends on the well-learned semantic parsing results. The balancing weights $\lambda_b$ and $\lambda_m$ in Eq. \ref{eq:distance} are all set $0.1$ by default. In experiments, we extend each point label to a square region with a size of 17 pixels to facilitate the training of semantic segmentation, as explained in Eq.~\ref{eq:pce_loss}. The shorter sizes of input images are resized to 600 and 800 on the VOC and COCO, respectively. On the VOC, we train 20 epochs and decay the learning rate with a factor of 0.1 after epoch 15. On the COCO, we follow the $1\times$ schedule to train 12 epochs and decay the learning rate after epoch 8.


\subsection{Ablation Studies}
In this section, we conduct analytic experiments on the VOC dataset to reveal the properties of the proposed method with point-level supervision.

\subsubsection{Instance Discrimination.}

We first conduct experiments to demonstrate the effectiveness of the proposed traversing distance-based instance discrimination approach. Tab.~\ref{tab:instance_discrimination} shows the ablation results of the distance measurement criteria in Eq.~\ref{eq:distance}. With only the semantic probabilities, the model achieves PQ 46.6\% on the VOC val set. The boundary criterion and the manifold criterion improve the baseline result to 48.5\% and 49.4\%, respectively. We noticed that this improvement is mainly due to the PQ\textsuperscript{th}, which is improved from 44.5\% to 46.5\% and 47.4\%, respectively. And the results of stuff classes are relatively similar, demonstrating that the low-level image cues and high-level instance-aware manifold can effectively help to identify different instances. Finally, with all the criteria, our approach achieves the final result of PQ 49.8\% and PQ\textsuperscript{th} 47.8\%.

\begin{table}[t]
\caption{Ablation studies for the proposed traversing distance-based instance discrimination. Results are reported on the VOC val set. $E^s$, $E^b$, and $E^m$ denote the criteria of semantic similarity, low-level boundary, and high-level manifold, respectively.}
\label{tab:instance_discrimination}
\centering
\tabcolsep=12pt
\begin{tabular}{ccc|ccc}
\hline

\hline
	$E^s$ & $E^b$ & $E^m$ & PQ & PQ\textsuperscript{th} & PQ\textsuperscript{st} \\
	\hline
	\checkmark & & & 46.6 & 44.5 & 89.1 \\
	\checkmark & \checkmark & & 48.5 & 46.5 & 89.3 \\
	\checkmark & & \checkmark & 	49.4 & 47.4 & 89.2 \\
	\checkmark & \checkmark & \checkmark & \textbf{49.8} & \textbf{47.8} & \textbf{89.5} \\
\hline

\hline
\end{tabular}

\end{table}

\subsubsection{Hyper-parameter Sensitivity.}

We conduct experiments to study the sensitivity of the hyper-parameters used in Eq.~\ref{eq:distance}. To save the search cost, we fix one parameter and adjust another. Results reported in Tab.~\ref{tab:hyperparameter} show that the scale of the additional criteria for the instance discrimination, i.e., the boundary and the feature manifold criteria, should be approximately one order of magnitude smaller than the semantic criterion. It is noteworthy that even larger values are not optimal, they still boost the baseline's performance from PQ 46.6\% to 48.3\% or higher, demonstrating the robustness of the proposed approach.

\begin{table}[t]
\caption{Influence of the hyper-parameters in computing the traversing cost. Results are reported on the VOC val set. We conduct experiments by fixing one hyper-parameter and alter another.}
\label{tab:hyperparameter}
\centering
\tabcolsep=14pt
\begin{tabular}{cc|ccc}
\hline

\hline
	$\lambda_b$ & $\lambda_m$ & PQ & PQ\textsuperscript{th} & PQ\textsuperscript{st} \\
	\hline
	0.0 & 0.1 & 49.4 & 47.4 & 89.2 \\	
	0.1 & 0.1 & \textbf{49.8} & \textbf{47.8} & \textbf{89.5} \\	
	0.5 & 0.1 & 48.3 & 46.3 & 89.1\\
	1.0 & 0.1 & 48.4 & 46.3 & 89.4 \\	
	\hline
	0.1 & 0.0 & 48.5 & 46.5 & 89.3 \\
	0.1 & 0.1 & \textbf{49.8} & \textbf{47.8} & \textbf{89.5} \\	
	0.1 & 0.5 &	49.3 & 47.3 & 89.2 \\	
	0.1 & 1.0 &	48.9 & 47.0 & 89.1 \\		
\hline

\hline
\end{tabular}
	
\end{table}

\subsubsection{Point Sampling Strategies.}
\label{sec:sampling_strategy}

We conduct experiments to study the influence of the position distribution bias of the point labels. In addition to the uniform sampling strategy, we also tried the center-biased and the border-biased sampling strategies. Specifically, we first compute the euclidean distance of each pixel to the centroid of the corresponding ground truth mask. Then, we build the probability density map according to the square of the euclidean distance. Finally, the points are sampled based on the normalized probability for each target to obtain border-biased labels. The center-biased labels are sampled in a similar way by reversing the probabilities. Results in Tab.~\ref{tab:point_sampling} show that the center-biased labels achieve the best performance, and the border-biased labels perform worst. While the SQ values of the three methods are relatively similar, the RQ of the border-biased strategy is much worse than the others, revealing that annotations near borders are harmful to discriminating different targets, while annotations at center regions provide more robust results. This phenomenon has positive meanings because center annotation accords with human intuition and is easier in practice.

\begin{table}[t]
\caption{Influence of the point sampling strategy. Results are reported on the VOC val set. ``Border'' and ``Center'' refer to strategies that prefer target border regions and center regions, respectively.}
\label{tab:point_sampling}
\centering
\tabcolsep=14pt
\begin{tabular}{l|cccc}
\hline

\hline
Method & mIoU & PQ & SQ & RQ \\
\hline
Border 	& 65.6 & 44.7 & 78.1 & 55.7 \\
Uniform & 67.5 & 49.8 & 78.4 & 62.0 \\
Center 	& \textbf{67.7} & \textbf{50.9} & \textbf{79.1} & \textbf{62.8} \\

\hline

\hline
\end{tabular}
\end{table}

\subsubsection{Semantic Segmentation Module.}
\label{sec:semantic_segmentation}
In this section, we conduct ablation experiments to study the semantic segmentation submodule. Results in Tab.~\ref{tab:semantic} demonstrate the low-level cues can effectively improve the segmentation performance from mIoU 62.2\% to mIoU 67.5\%. The improvement of the semantic segmentation quality not only improves the quality of the panoptic masks, i.e., SQ is improved from 74.3\% to 78.4\%, but also benefits the localization of the targets, i.e., a significant improvement of the RQ from 54.0\% to 62.0\%. We conjecture the reason is that more accurate semantic probabilities provide more reliable instance discrimination cues for Eq.~\ref{eq:distance} to distinguish different targets.

\begin{table}[t]
\caption{Ablation study of the semantic segmentation submodule. Results are reported on the VOC val set.}
\label{tab:semantic}
\centering
\tabcolsep=14pt
\begin{tabular}{r|cccc}
\hline

\hline
Method & mIoU & PQ & SQ & RQ \\
\hline
w/o $\mathcal{L}_{col}$ & 62.2 & 41.3 & 74.3 & 54.0 \\
w/~ $\mathcal{L}_{col}$  & 67.5 & 49.8 & 78.4 & 62.0 \\
\hline

\hline
\end{tabular}
\end{table}

\subsection{Comparison with Related Works}

We compare our approach with the related works in Tab.~\ref{tab:sota}.
Compared to the JTSM~\cite{wsss:jtsm} that uses image-level labels, our approach achieves significant improvement with the help of point-level labels, which improves the PQ with +10.8\% on the VOC, and +24.0\% on the COCO. It demonstrates that point-level labels are promising in addressing panoptic segmentation problems. As pointed out by Bearman et al.~\cite{wsss:whats_point}, the point-level labels only cost marginally above image-level labels. For example, on the VOC dataset, image labels cost on average 20.0~sec/img, while point labels cost 22.1~sec/img, where the difference is marginal compared with the full labels' 239.7~sec/img.  
Compared to the PanopticFCN~\cite{panoptic_fcn} using ten points to connect polygons for training, our approach achieves competitive performance when using only a single point as annotation, which is +1.8\% on the VOC and -1.9\% on the COCO in respect to the PQ. Note that these comparable results are achieved by using only $1/10$ of the annotations of the PanopticFCN. When increasing the annotation to ten points per target, our approach achieves +8.6\% and +1.9\% improvements on the VOC and the COCO datasets compared with the PanopticFCN, demonstrating the scalability of our approach in utilizing more points per target.

\begin{figure}[t]
\centering
\includegraphics[width=0.84\textwidth]{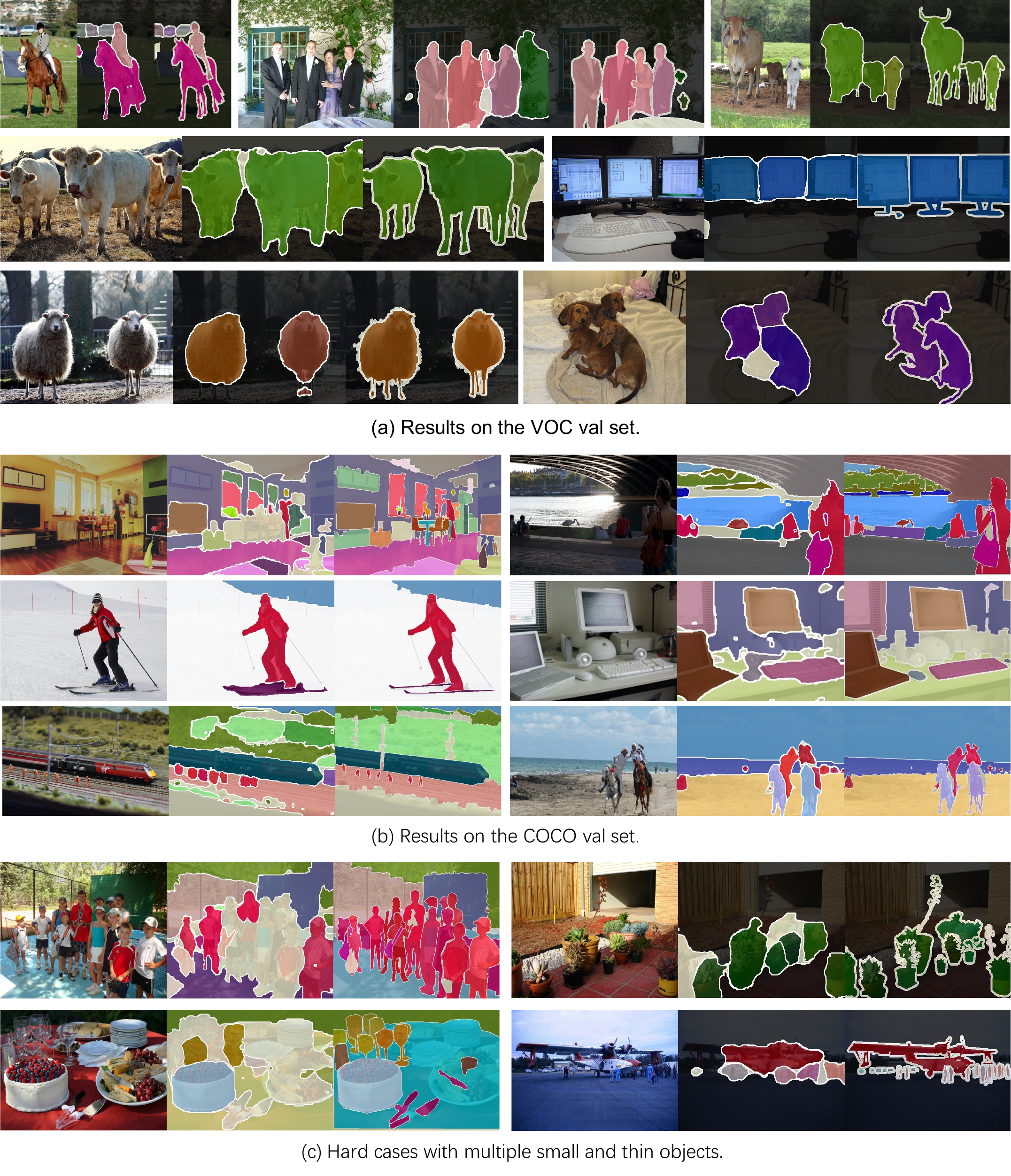}
\caption{Visualization of the panoptic segmentation results. The models are trained with a \textit{single point} per target as annotation. Each group from left to right are the input, prediction, and ground truth, respectively. Best viewed in color.}
\label{fig:results}
\end{figure}

To help qualitatively study the results, we visualize the predictions on the val set of VOC and COCO. Results in Fig.~\ref{fig:results} show that our approach performs generally well in handling scenes with complex multiple instances and classes. We also show the results for hard examples, which contain extremely many instances with small scales. In these cases, some instances are missing in the prediction. Improving the performance with these small and thin objects when only accessing point-level labels may be a potential direction in future studies.

\begin{table}[t]
\caption{Comparison with related works on the VOC and the COCO datasets. Results are reported on the COCO val set and the VOC val set. $\mathcal{M}$ mask annotation for fully-supervised learning, $\mathcal{B}$ bounding-box level supervision, $\mathcal{I}$ image-level supervision, $\mathcal{P}$ the proposed point-level supervision, $\mathcal{P}_{10}$ point-level supervision with 10 points per target.}
\label{tab:sota}
\centering
\tabcolsep=2.5pt
\begin{tabular}{lcc ccc ccc}
\hline

\hline
\multirow{2}{*}{Method} & \multirow{2}{*}{Backbone} & \multirow{2}{*}{Supervision} & \multicolumn{3}{c}{COCO} & \multicolumn{3}{c}{VOC} \\
\cline{4-9}
&&& PQ & PQ\textsuperscript{th} & PQ\textsuperscript{st} & PQ & PQ\textsuperscript{th} & PQ\textsuperscript{st} \\
\hline
Panoptic FPN \cite{panoptic_fpn}		& R50 & $\mathcal{M}$ & 41.5 & 48.3 & 31.2 & 65.7 & 64.5 & 90.8 \\
K-Net 		\cite{knet}		& R50 & $\mathcal{M}$ & 47.1 & 51.7 & 40.3 & - & - & - \\
MaskFormer \cite{maskformer}			& R50 & $\mathcal{M}$ & 46.5 & 51.0 & 39.8 & - & - & - \\
Panoptic SegFormer \cite{panoptic_segformer} 	& R50 & $\mathcal{M}$ & 48.0 & 52.3 & 41.5 & 67.9 & 66.6 & 92.7 \\
\hline
Li et.al.	\cite{wsss:wsps}		& R101 & $\mathcal{B}+\mathcal{I}$ & - & - & - & 59.0 & - & - \\
JTSM 	\cite{wsss:jtsm}			& R18-WS & $\mathcal{I}$ & 5.3 & 8.4 & 0.7 & 39.0 & 37.1 & 77.7 \\
PanopticFCN \cite{panoptic_fcn}		& R50 & $\mathcal{P}_{10}$ & 31.2 & 35.7 & 24.3 & 48.0 & 46.2 & 85.2 \\
\hline
Ours 				& R50 & $\mathcal{P}$ & 29.3 & 29.3 & 29.4 & 49.8 & 47.8 & 89.5 \\
Ours 				& R50 & $\mathcal{P}_{10}$ & 33.1 & 33.6 & 32.2 & 56.6 & 54.8 & 91.4 \\

\hline

\hline

\end{tabular}
	
\end{table}

\section{Conclusions}

In this paper, we propose a new paradigm for weakly-supervised panoptic segmentation using a single point label for each target. To tackle this problem, we propose a principled approach that generates panoptic pseudo-masks by solving the minimization problem of pixel-to-point traversing costs, which integrates semantic similarity, low-level texture cues, and high-level knowledge to distinguish different targets. We demonstrate its effectiveness and study the influence of point labels through analytical experiments. Besides, we achieve new state-of-the-art performance with point-level labels on the VOC and COCO datasets.

\paragraph{\textbf{Acknowledgments}} This work was supported in part by 
the Major Project for New Generation of AI (No.2018AAA0100400),
the National Natural Science Foundation of China (No. 61836014, No. U21B2042, No. 62072457, No. 62006231).

\clearpage
%
%
\bibliographystyle{splncs04}
\bibliography{bib/wsss,bib/data,bib/net,bib/sal,bib/seg,bib/selfsup,bib/transformer,bib/others}

\begin{thebibliography}{10}
\providecommand{\url}[1]{\texttt{#1}}
\providecommand{\urlprefix}{URL }
\providecommand{\doi}[1]{https://doi.org/#1}

\bibitem{wsss:irnet}
Ahn, J., Cho, S., Kwak, S.: Weakly supervised learning of instance segmentation
  with inter-pixel relations. In: Proceedings of the IEEE/CVF Conference on
  Computer Vision and Pattern Recognition. pp. 2209--2218 (2019)

\bibitem{wsss:whats_point}
Bearman, A., Russakovsky, O., Ferrari, V., Fei-Fei, L.: What's the point:
  Semantic segmentation with point supervision. In: Proceedings of the European
  conference on computer vision. pp. 549--565. Springer (2016)

\bibitem{detr}
Carion, N., Massa, F., Synnaeve, G., Usunier, N., Kirillov, A., Zagoruyko, S.:
  End-to-end object detection with transformers. In: Proceedings of the
  European conference on computer vision. pp. 213--229. Springer (2020)

\bibitem{survey:wsss}
Chan, L., Hosseini, M.S., Plataniotis, K.N.: A comprehensive analysis of
  weakly-supervised semantic segmentation in different image domains.
  International Journal of Computer Vision  \textbf{129}(2),  361--384 (2021)

\bibitem{chen2021points}
Chen, L., Yang, T., Zhang, X., Zhang, W., Sun, J.: Points as queries: Weakly
  semi-supervised object detection by points. In: Proceedings of the IEEE/CVF
  Conference on Computer Vision and Pattern Recognition. pp. 8823--8832 (2021)

\bibitem{selfsup:simclr}
Chen, T., Kornblith, S., Norouzi, M., Hinton, G.: A simple framework for
  contrastive learning of visual representations. In: International conference
  on machine learning. pp. 1597--1607. PMLR (2020)

\bibitem{wsss:psis}
Cheng, B., Parkhi, O., Kirillov, A.: Pointly-supervised instance segmentation.
  In: Proceedings of the IEEE/CVF Conference on Computer Vision and Pattern
  Recognition. pp. 2617--2626 (2022)

\bibitem{maskformer}
Cheng, B., Schwing, A., Kirillov, A.: Per-pixel classification is not all you
  need for semantic segmentation. Advances in Neural Information Processing
  Systems  \textbf{34} (2021)

\bibitem{wsss:boxsup}
Dai, J., He, K., Sun, J.: Boxsup: Exploiting bounding boxes to supervise
  convolutional networks for semantic segmentation. In: Proceedings of the IEEE
  International Conference on Computer Vision. pp. 1635--1643 (2015)

\bibitem{jsis}
De~Geus, D., Meletis, P., Dubbelman, G.: Panoptic segmentation with a joint
  semantic and instance segmentation network. arXiv preprint arXiv:1809.02110
  (2018)

\bibitem{bert}
Devlin, J., Chang, M.W., Lee, K., Toutanova, K.: Bert: Pre-training of deep
  bidirectional transformers for language understanding. arXiv preprint
  arXiv:1810.04805  (2018)

\bibitem{vit}
Dosovitskiy, A., Beyer, L., Kolesnikov, A., Weissenborn, D., Zhai, X.,
  Unterthiner, T., Dehghani, M., Minderer, M., Heigold, G., Gelly, S.,
  Uszkoreit, J., Houlsby, N.: An image is worth 16x16 words: Transformers for
  image recognition at scale. ICLR  (2021)

\bibitem{data:voc}
Everingham, M., Van~Gool, L., Williams, C.K., Winn, J., Zisserman, A.: The
  pascal visual object classes (voc) challenge. International Journal of
  Computer Vision  \textbf{88}(2),  303--338 (2010)

\bibitem{wsss:icd}
Fan, J., Zhang, Z., Song, C., Tan, T.: Learning integral objects with
  intra-class discriminator for weakly-supervised semantic segmentation. In:
  Proceedings of the IEEE/CVF Conference on Computer Vision and Pattern
  Recognition. pp. 4283--4292 (2020)

\bibitem{wsss:multi_seed}
Fan, J., Zhang, Z., Tan, T.: Employing multi-estimations for weakly-supervised
  semantic segmentation. In: Proceedings of the European Conference on Computer
  Vision (2020)

\bibitem{wsss:cian}
Fan, J., Zhang, Z., Tan, T., Song, C., Xiao, J.: {CIAN}: Cross-image affinity
  net for weakly supervised semantic segmentation. In: Proceedings of the AAAI
  Conference on Artificial Intelligence. vol.~34, pp. 10762--10769 (2020)

\bibitem{wsss:aisi}
Fan, R., Hou, Q., Cheng, M.M., Yu, G., Martin, R.R., Hu, S.M.: Associating
  inter-image salient instances for weakly supervised semantic segmentation.
  In: Proceedings of the European Conference on Computer Vision. pp. 367--383
  (2018)

\bibitem{data:voc_extra}
Hariharan, B., Arbel{\'a}ez, P., Bourdev, L., Maji, S., Malik, J.: Semantic
  contours from inverse detectors. In: Proceedings of the International
  Conference on Computer Vision. pp. 991--998. IEEE (2011)

\bibitem{selfsup:moco}
He, K., Fan, H., Wu, Y., Xie, S., Girshick, R.: Momentum contrast for
  unsupervised visual representation learning. In: Proceedings of the IEEE/CVF
  Conference on Computer Vision and Pattern Recognition. pp. 9729--9738 (2020)

\bibitem{net:resnet_v1}
He, K., Zhang, X., Ren, S., Sun, J.: Deep residual learning for image
  recognition. In: Proceedings of the IEEE Conference on Computer Vision and
  Pattern Recognition. pp. 770--778 (2016)

\bibitem{wsss:sdi_box}
Khoreva, A., Benenson, R., Hosang, J., Hein, M., Schiele, B.: Simple does it:
  Weakly supervised instance and semantic segmentation. In: Proceedings of the
  IEEE conference on computer vision and pattern recognition. pp. 876--885
  (2017)

\bibitem{panoptic_fpn}
Kirillov, A., Girshick, R., He, K., Doll{\'a}r, P.: Panoptic feature pyramid
  networks. In: Proceedings of the IEEE/CVF Conference on Computer Vision and
  Pattern Recognition. pp. 6399--6408 (2019)

\bibitem{panoptic_segmentation}
Kirillov, A., He, K., Girshick, R., Rother, C., Doll{\'a}r, P.: Panoptic
  segmentation. In: Proceedings of the IEEE/CVF Conference on Computer Vision
  and Pattern Recognition. pp. 9404--9413 (2019)

\bibitem{wsss:sec}
Kolesnikov, A., Lampert, C.H.: Seed, expand and constrain: Three principles for
  weakly-supervised image segmentation. In: Proceedings of the European
  Conference on Computer Vision. pp. 695--711. Springer (2016)

\bibitem{alg:crf}
Kr{\"a}henb{\"u}hl, P., Koltun, V.: Efficient inference in fully connected crfs
  with gaussian edge potentials. In: Advances in Neural Information Processing
  Systems. pp. 109--117 (2011)

\bibitem{wsss:ficklenet}
Lee, J., Kim, E., Lee, S., Lee, J., Yoon, S.: Ficklenet: Weakly and
  semi-supervised semantic image segmentation using stochastic inference. In:
  Proceedings of the IEEE Conference on Computer Vision and Pattern
  Recognition. pp. 5267--5276 (2019)

\bibitem{wsss:wsps}
Li, Q., Arnab, A., Torr, P.H.: Weakly-and semi-supervised panoptic
  segmentation. In: Proceedings of the European Conference on Computer Vision.
  pp. 102--118 (2018)

\bibitem{unify_panoptic}
Li, Q., Qi, X., Torr, P.H.: Unifying training and inference for panoptic
  segmentation. In: Proceedings of the IEEE/CVF Conference on Computer Vision
  and Pattern Recognition. pp. 13320--13328 (2020)

\bibitem{panoptic_fcn}
Li, Y., Zhao, H., Qi, X., Chen, Y., Qi, L., Wang, L., Li, Z., Sun, J., Jia, J.:
  Fully convolutional networks for panoptic segmentation with point-based
  supervision. arXiv preprint arXiv:2108.07682  (2021)

\bibitem{panoptic_segformer}
Li, Z., Wang, W., Xie, E., Yu, Z., Anandkumar, A., Alvarez, J.M., Luo, P., Lu,
  T.: Panoptic segformer: Delving deeper into panoptic segmentation with
  transformers. In: Proceedings of the IEEE/CVF Conference on Computer Vision
  and Pattern Recognition. pp. 1280--1289 (2022)

\bibitem{wsss:scribblesup}
Lin, D., Dai, J., Jia, J., He, K., Sun, J.: Scribblesup: Scribble-supervised
  convolutional networks for semantic segmentation. In: Proceedings of the IEEE
  Conference on Computer Vision and Pattern Recognition. pp. 3159--3167 (2016)

\bibitem{data:coco}
Lin, T.Y., Maire, M., Belongie, S., Bourdev, L., Girshick, R., Hays, J.,
  Perona, P., Ramanan, D., Zitnick, C.L., Doll{\'a}r, P.: Microsoft coco:
  Common objects in context. In: Proceedings of the European Conference on
  Computer Vision. pp. 740--755. Springer (2014)

\bibitem{crowd_counting2}
Liu, Y., Xu, D., Ren, S., Wu, H., Cai, H., He, S.: Fine-grained domain adaptive
  crowd counting via point-derived segmentation. arXiv preprint
  arXiv:2108.02980  (2021)

\bibitem{swin}
Liu, Z., Lin, Y., Cao, Y., Hu, H., Wei, Y., Zhang, Z., Lin, S., Guo, B.: Swin
  transformer: Hierarchical vision transformer using shifted windows. In:
  Proceedings of the IEEE/CVF International Conference on Computer Vision. pp.
  10012--10022 (2021)

\bibitem{deep_extreme_cut}
Maninis, K.K., Caelles, S., Pont-Tuset, J., Van~Gool, L.: Deep extreme cut:
  From extreme points to object segmentation. In: Proceedings of the IEEE
  Conference on Computer Vision and Pattern Recognition. pp. 616--625 (2018)

\bibitem{wsss:ccnn}
Pathak, D., Kr\"{a}henb\"{u}hl, P., Darrell, T.: Constrained convolutional
  neural networks for weakly supervised segmentation. In: Proceedings of the
  IEEE International Conference on Computer Vision. pp. 1796--1804 (2015)

\bibitem{peng2021spts}
Peng, D., Wang, X., Liu, Y., Zhang, J., Huang, M., Lai, S., Zhu, S., Li, J.,
  Lin, D., Shen, C., et~al.: Spts: Single-point text spotting. arXiv preprint
  arXiv:2112.07917  (2021)

\bibitem{wsss:pdml}
Qian, R., Wei, Y., Shi, H., Li, J., Liu, J., Huang, T.: Weakly supervised scene
  parsing with point-based distance metric learning. In: Proceedings of the
  AAAI Conference on Artificial Intelligence. vol.~33, pp. 8843--8850 (2019)

\bibitem{gpt}
Radford, A., Wu, J., Child, R., Luan, D., Amodei, D., Sutskever, I., et~al.:
  Language models are unsupervised multitask learners. OpenAI blog
  \textbf{1}(8), ~9 (2019)

\bibitem{wsss:grad_cam}
Selvaraju, R.R., Cogswell, M., Das, A., Vedantam, R., Parikh, D., Batra, D.:
  Grad-cam: Visual explanations from deep networks via gradient-based
  localization. In: Proceedings of the IEEE International Conference on
  Computer Vision. pp. 618--626 (2017)

\bibitem{wsss:jtsm}
Shen, Y., Cao, L., Chen, Z., Lian, F., Zhang, B., Su, C., Wu, Y., Huang, F.,
  Ji, R.: Toward joint thing-and-stuff mining for weakly supervised panoptic
  segmentation. In: Proceedings of the IEEE/CVF Conference on Computer Vision
  and Pattern Recognition. pp. 16694--16705 (2021)

\bibitem{iterative_iterative}
Sofiiuk, K., Petrov, I.A., Konushin, A.: Reviving iterative training with mask
  guidance for interactive segmentation. arXiv preprint arXiv:2102.06583
  (2021)

\bibitem{wsss:frgbox}
Song, C., Huang, Y., Ouyang, W., Wang, L.: Box-driven class-wise region masking
  and filling rate guided loss for weakly supervised semantic segmentation. In:
  Proceedings of the IEEE Conference on Computer Vision and Pattern
  Recognition. pp. 3136--3145 (2019)

\bibitem{segmenter}
Strudel, R., Garcia, R., Laptev, I., Schmid, C.: Segmenter: Transformer for
  semantic segmentation. In: Proceedings of the IEEE/CVF International
  Conference on Computer Vision. pp. 7262--7272 (2021)

\bibitem{wsss:boxinst}
Tian, Z., Shen, C., Wang, X., Chen, H.: Boxinst: High-performance instance
  segmentation with box annotations. In: Proceedings of the IEEE/CVF Conference
  on Computer Vision and Pattern Recognition. pp. 5443--5452 (2021)

\bibitem{transformer}
Vaswani, A., Shazeer, N., Parmar, N., Uszkoreit, J., Jones, L., Gomez, A.N.,
  Kaiser, {\L}., Polosukhin, I.: Attention is all you need. Advances in neural
  information processing systems  \textbf{30} (2017)

\bibitem{wsss:scribblerw}
Vernaza, P., Chandraker, M.: Learning random-walk label propagation for
  weakly-supervised semantic segmentation. In: Proceedings of the IEEE
  Conference on Computer Vision and Pattern Recognition. vol.~3, p.~3 (2017)

\bibitem{wsss:mdc}
Wei, Y., Xiao, H., Shi, H., Jie, Z., Feng, J., Huang, T.S.: Revisiting dilated
  convolution: A simple approach for weakly- and semi-supervised semantic
  segmentation. In: Proceedings of the IEEE Conference on Computer Vision and
  Pattern Recognition. pp. 7268--7277 (2018)

\bibitem{auto_panoptic}
Wu, Y., Zhang, G., Xu, H., Liang, X., Lin, L.: Auto-panoptic: Cooperative
  multi-component architecture search for panoptic segmentation. Advances in
  Neural Information Processing Systems  \textbf{33},  20508--20519 (2020)

\bibitem{crowd_counting}
Zand, M., Damirchi, H., Farley, A., Molahasani, M., Greenspan, M., Etemad, A.:
  Multiscale crowd counting and localization by multitask point supervision.
  In: ICASSP 2022-2022 IEEE International Conference on Acoustics, Speech and
  Signal Processing (ICASSP). pp. 1820--1824. IEEE (2022)

\bibitem{survey:wsod}
Zhang, D., Han, J., Cheng, G., Yang, M.H.: Weakly supervised object
  localization and detection: a survey. IEEE Transactions on Pattern Analysis
  and Machine Intelligence  (2021)

\bibitem{survey:wsss2}
Zhang, M., Zhou, Y., Zhao, J., Man, Y., Liu, B., Yao, R.: A survey of semi-and
  weakly supervised semantic segmentation of images. Artificial Intelligence
  Review  \textbf{53}(6),  4259--4288 (2020)

\bibitem{iog}
Zhang, S., Liew, J.H., Wei, Y., Wei, S., Zhao, Y.: Interactive object
  segmentation with inside-outside guidance. In: Proceedings of the IEEE/CVF
  Conference on Computer Vision and Pattern Recognition. pp. 12234--12244
  (2020)

\bibitem{knet}
Zhang, W., Pang, J., Chen, K., Loy, C.C.: K-net: Towards unified image
  segmentation. Advances in Neural Information Processing Systems  \textbf{34}
  (2021)

\bibitem{wsss:cam}
Zhou, B., Khosla, A., Lapedriza, A., Oliva, A., Torralba, A.: Learning deep
  features for discriminative localization. In: Proceedings of the IEEE
  Conference on Computer Vision and Pattern Recognition. pp. 2921--2929 (2016)

\bibitem{deformable_detr}
Zhu, X., Su, W., Lu, L., Li, B., Wang, X., Dai, J.: Deformable detr: Deformable
  transformers for end-to-end object detection. In: Proceedings of the
  International Conference on Learning Representations (2021)

\end{thebibliography}

\end{document}